\documentclass[pdflatex,sn-vancouver-num]{sn-jnl}% Vancouver Numbered Reference Style
\usepackage{graphicx}%
\usepackage{multirow}%
\usepackage{amsmath,amssymb,amsfonts}%
\usepackage{amsthm}%
\usepackage{mathrsfs}%
\usepackage[title]{appendix}%
\usepackage{xcolor}%
\usepackage{textcomp}%
\usepackage{manyfoot}%
\usepackage{booktabs}%
\usepackage{algorithm}%
\usepackage{algorithmicx}%
\usepackage{algpseudocode}%
\usepackage{listings}%
\usepackage{graphicx}
\usepackage{amsmath}
\usepackage{amssymb}
\usepackage{booktabs}
\usepackage{tikz}
\usepackage{amsfonts}
\usepackage{array} 
\usepackage{amsmath}
\usepackage{xcolor}         % colors
\usepackage{graphicx}
\usepackage{amsmath}
\usepackage{amssymb}
\usepackage{booktabs}
\usepackage{multirow}
\usepackage{pifont}
\usepackage{pifont} % for tick/cross
\usepackage{amsmath}
\theoremstyle{thmstyleone}%
\theoremstyle{thmstyletwo}%

\theoremstyle{thmstylethree}%

\begin{document}

\title[Article Title]{Cranio-Diff: Diffusion-based Cross-domain Craniofacial Reconstruction with 2D X-ray Skull}

%=============================================================%%
% GivenName	-> \fnm{Joergen W.}
% Particle	-> \spfx{van der} -> surname prefix
% FamilyName	-> \sur{Ploeg}
% Suffix	-> \sfx{IV}
% \author*[1,2]{\fnm{Joergen W.} \spfx{van der} \sur{Ploeg} 
%  \sfx{IV}}\email{iauthor@gmail.com}
%=============================================================%%

\author*[1]{\fnm{Ravi Shankar} \sur{Prasad}}\email{d23033@students.iitmandi.ac.in, dineshsingh@iitmandi.ac.in}

\author[1]{\fnm{Naresh} \sur{Gurjar}}\email{ng4955607@gmail.com}

\author[1]{\fnm{Shashank} \sur{Baghel}}\email{s25038@students.iitmandi.ac.in}
\author[1]{\fnm{Chirag} }\email{t25089@students.iitmandi.ac.in}
\author[1]{\fnm{Dinesh} \sur{Singh}}

\affil*[1]{\orgdiv{Visual Intelligence and Machine Learning Group, School of Computing and Electrical Engineering}, \orgname{Indian Institute of Technology Mandi}, \orgaddress{\street{Kamand}, \city{Mandi}, \postcode{175075}, \state{Himachal Pradesh}, \country{India}}}

% \affil[2]{\orgdiv{University Teaching Department}, \orgname{Swami Vivekanand Technical University (CSVTU)}, \orgaddress{ \city{Bhilai}, \postcode{10587}, \state{Chhattisgarh}, \country{India}}}

\abstract{The identification of unknown human remains is a persistent humanitarian and forensic challenge. In forensic investigations, it becomes even more challenging when common biometric DNA identifications, such as soft-tissue or hair strands are not present, as they degrade with time. To narrow down the process of identification, few attempts have been made to identify an individual by reconstructing a face from the given skeletal remains ( i.e., skull) using generative AI models, such as CycleGAN, Pix2Pix, and diffusion models. However, they fail to effectively capture semantic information in craniofacial reconstruction, due to a mismatch in the alignment of structural information across modalities. To address this issue, we propose \textit{Cranio-Diff}, a diffusion-based framework for craniofacial reconstruction from 2D X-ray skull images. The proposed approach integrates skull-conditioned structural guidance through ControlNet with biometric text prompts to generate a face. The proposed \textit{Cranio-Diff} method is evaluated on skull-face dataset obtained from X-ray scans of 120 subjects in lateral and frontal views. To enable controlled evaluation, each face image is artificially synthesized across three age groups (25, 45, 65) and three BMI variations $\pm10\%$, yielding 4320 paired samples. This dataset is also curated across different facial modalities ( RGB and sketch) with different appearance conditions (hair and without hair). To the best of our knowledge, this is the only X-ray-face dataset with this magnitude. Extensive experiments showed that the proposed method achieves improved performance over the recent existing approaches in both generated image quality and retrieval task. Finally, to evaluate the performance of our proposed method, we have evaluated the quality of the generated image using FID, SSIM, LPIPS score. Additionally, retrieval performance is evaluated using recall@k, mAP@k and MRR@k. Obtained experimental results demonstrate that the proposed method can be used as an alternate tool in providing aid in forensic investigations.

}


\maketitle

\section{Introduction}
\label{sec:intro}

Identifying a deceased individual from skeletal remains is a challenging task, particularly in the absence of common biometric information such as soft tissue, hair strands, etc. Forensic Craniofacial Reconstruction (FCR) is one of the techniques that is widely used in forensic and anthropological applications. Traditionally, FCR is done by manually overlapping clay onto the skull with the help of soft-tissue depth information. But this manual process is time-taking, requires anatomical expertise and is highly subjective. To bypass this manual FCR, a few studies were conducted on computerized methods~\cite{claes2010computerized,huang2011weighted,guerra2025international}. However, due to complex structure of craniofacial features, these methods were not effective in mapping skull-to-face features.

In the direction of craniofacial identification, direct skull-to-face matching~\cite{prasad2026cross,prasad2026spot,prasad2025cranio} can be seen as an alternate solution but this task is overburdened by model to map the representation of skull with the face, also it does not aid to the deceased individual's family and friend to visually identify them. The recent methods in FCR perform skull-to-face reconstruction~\cite{li2022cr, prasad2025fcr} using different generative AI models (i.e., GANs). Although these generative models have claimed good performance in terms of mapping skull structure to the facial appearance. But, due to very limited dataset and absence of any public benchmark dataset in this domain, these generative models often fail to generalize in preserving structural identity of individuals under different conditions such as age and gender variations. In the recent years, diffusion-based models~\cite{zhang2025cr, zhang2025iccr} have emerged as state-of-the-art methods in generating improved facial images from 3D CT skull images with improved training stability. Furthermore, the introduction of multimodal text-conditioned diffusion models~\cite{zhao2023uni} has enabled flexible and controllable image synthesis by incorporating semantic guidance through textual prompts. Also, these methods primarily focus on visual appearance and lack explicit mechanisms to enforce anatomical consistency. However, existing diffusion-based approaches for craniofacial reconstruction conducted on 3D CT scans, as collection of 3D CT scan has following limitations. First, high radiation exposure. Second, need high cost device and setup. Third, high compute and processing time. 
% Hence, craniofacial identification methods are not the final identification but, rather it aid the recognition process of deceased individuals.

Despite advances in these generative models, we have found three key limitations that remain in existing craniofacial reconstruction methods. \textbf{First}, there is a lack of multimodal conditioning that integrates both structural (skull/X-ray) and semantic (textual) information. \textbf{Second}, absence of explicit geometric constraints to enforce anatomical consistency. \textbf{Third}, limited exploration of cross-modal reconstruction from 2D X-ray images.

Hence, to address these challenges, we propose \textit{Cranio-Diff}, a diffusion-based framework for forensic skull-to-face reconstruction from skull images. The proposed 
framework integrates a trainable ControlNet-guided~\cite{zhao2023uni} Stable Diffusion v1.5~\cite{rombach2022high} backbone, which has been fine-tuned for photorealistic facial synthesis (Realistic Vision v5.1\footnote{\url{https://huggingface.co/SG161222/Realistic_Vision_V5.1_noVAE}}). This enables to generate photorealistic facial reconstructions directly from skull anatomy. Also, few studies were conducted on 3D skull and 3D face due unavailability of public benchmark dataset and these 3D scans requires large amount of resources to collect and process the data. Hence, this study benefits from the use of 2D skull and 2D face image datasets, which significantly reduce redundant information compared to 3D and are also easier to collect than 3D. To the best of our knowledge, no comprehensive automated craniofacial
reconstruction techniques have been proposed to directly recreate high quality 2D images of the face from the 2D X-ray scan data with variation of age and body mass index.
\begin{figure}[!t]
        \centering
        \includegraphics[width =\linewidth,keepaspectratio,trim={1cm 2cm 3cm 5cm},clip]{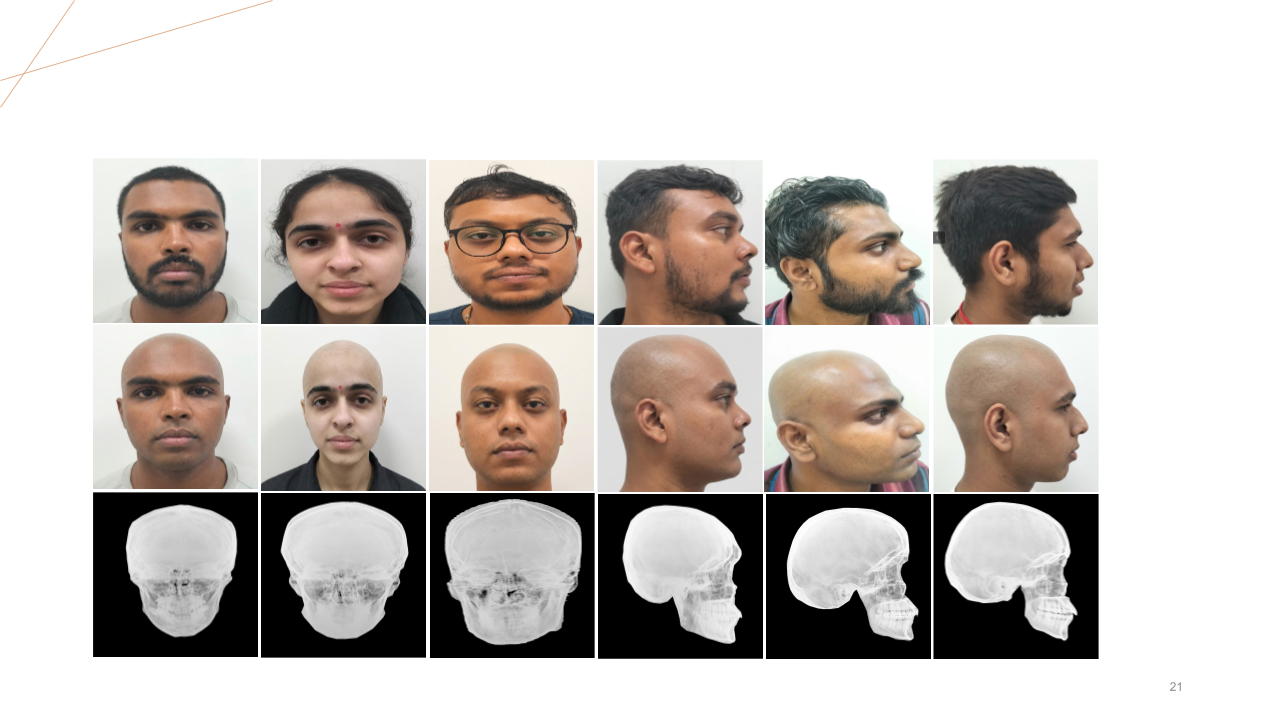}
        \caption{Face-skull dataset with cleaned and ground truth face images.}
        \label{fig:s2f_dataset}
    \end{figure}

The major contributions of this research are as follows:
\begin{itemize}
\item We propose \textit{Cranio-Diff}, a diffusion-based framework for craniofacial reconstruction from multi-view 2D X-ray images with variation of age and Body mass index.
\item We introduce a multi-modal conditioning strategy that integrates structural information and textual prompts for anatomically consistent and controllable face generation.
\item We have conducted extensive experiments on multimodal skull-to-face generation and evaluated their performance using different metrics such as FID, SSIM, LPIPS score, demonstrating the effectiveness of the proposed approach.

\item We synthesize proposed dataset with controlled age (25, 45 and 65) and BMI (-10\%, baseline, +10\%) variations, and also curated the dataset across four facial modalities (RGB face with hair, hair-removed RGB face, sketch face with hair, and hair-removed sketch face), supporting diverse forensic matching scenarios and enabling systematic evaluation of skull-face association under realistic appearance changes.

\item Additionally, we have conducted experiments on retrieval of face from the gallery of face images for the given query (i.e., reconstructed face image) from the skull using three face recognition backbones (i.e., FaceNet, ArcFace and VGGFace) and evaluated their retrieval performance using Recall@k, mean Average Precision (mAP@k) and Mean Reciprocal Rank (MRR@k) metrics.
\vspace{0.2cm}

The rest of the paper is organized as follows. In Section \ref{related work},
we introduce the latest research on Craniofacial Identification using reconstruction. In Section \ref{proposed dataset}, we present the
details of the dataset proposed in this paper. In Section \ref{methods},
we introduce the methodology for Craniofacial reconstruction. Implementation details, experimental results and a detailed analysis are
discussed in Section \ref{experiments}. The conclusion is drawn in Section \ref{conclusion}, followed by Declaration and Acknowledgment in Section \ref{declaration} and Section \ref{anknowledgment}, respectively.
\end{itemize}

\section{Related work}\label{related work}

\textbf{Traditional Craniofacial Reconstruction.}  Craniofacial reconstruction has traditionally been performed manually by forensic experts, who superimpose facial information over a skull using anatomical guidelines and tissue depth markers information~\cite{wilkinson2010facial,hwang2012facial}. These approaches are labour-intensive, require significant expertise, and lack scalability. Moreover, they offer limited evaluation due to the absence of reliable ground truth data.
To address these limitations faced by traditional methods, early computational methods use statistical learning techniques such as least squares support vector regression~\cite{li2014craniofacial}, latent root regression~\cite{berar2011craniofacial}, and partial least squares regression~\cite{jia2021craniofacial}. Although these approaches are used to model global relationships between skull and facial features which are heavily on predefined landmarks and struggle to capture fine-grained local details. Additionally, their performance is constrained by limited training data and weak generalization across populations.

\textbf{Deep learning-based approaches.} With the recent advancement in deep learning models and their application in the field of generative AI, the task of image generations has been significantly improved. GAN-based frameworks~\cite{goodfellow2014generative}, particularly Pix2pix (cGAN)~\cite{isola2017image} and cycleGAN~\cite{zhu2017unpaired}, have been widely used for cross-domain image-to-image synthesis due to their ability to learn mappings between unpaired domains. However, these approaches in craniofacial reconstruction~\cite{li2022cr,prasad2025fcr} often struggle to effectively preserve structural attributes and are difficult to leverage multimodal information (like text and images), especially in scenarios involving large geometric and structural variations between skulls and facial appearance. Hence, GAN-based methods remain limited by their training instability and tendency to produce blurred or structurally inconsistent results.
% \begin{figure*}[!ht]
%         \centering
%         \includegraphics[width =\linewidth,keepaspectratio,trim={0cm 3cm 0cm 1cm},clip]{Figure/diverse_s2f.png}
%         \caption{Sample images of curated dataset, showing variation across age (25, 45, 65) and BMI (-10\%, baseline, +10\%) for frontal and lateral view face image.}
%         \label{fig:diverse_s2f}
%     \end{figure*}

\begin{figure*}[!ht]
        \centering
        \includegraphics[width =\linewidth,keepaspectratio,trim={0cm 0cm 0cm 1cm},clip]{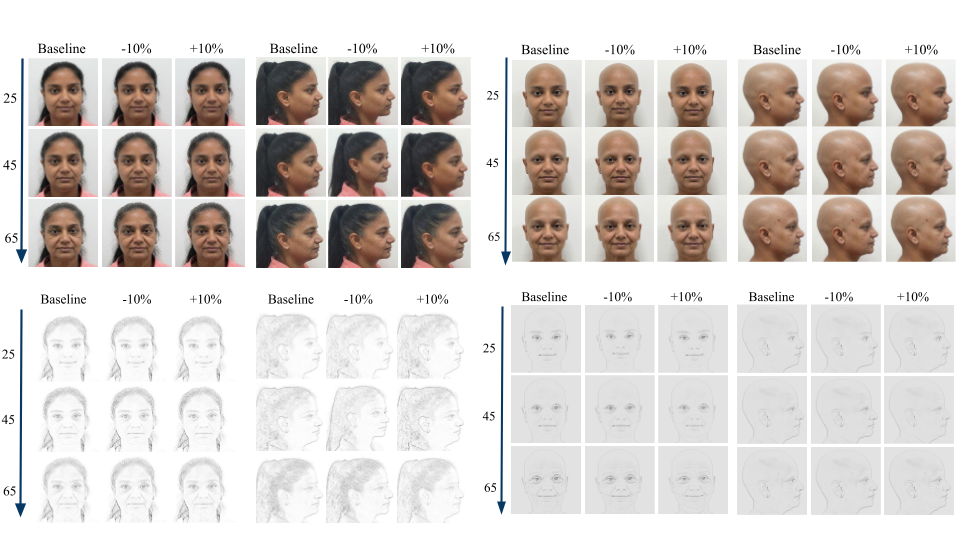}
        \caption{Sample images of Subject ID 21 from the curated dataset across different modalities ( RGB and sketch), with different facial appearances ( with and without hair) showing variation across age (25, 45, 65) and BMI (-10\%, baseline, +10\%) for frontal and lateral view face image.}
        \label{fig:full_s2f}
    \end{figure*}
% \begin{figure}[!ht]
%         \centering
%         \includegraphics[width =0.5\linewidth,keepaspectratio,trim={7cm 1cm 8cm 1cm},clip]{Figure/pairs.pdf}
%         \caption{Example from CranioBench dataset. Here, face images (A, B, C, D) with text prompts are shown across different modalities and variations with their corresponding skull image in E.}
%         \label{fig:sample}
%     \end{figure}

Later on very few studies were done on diffusion-based approaches~\cite{zhang2025cr,zhang2025iccr} to reconstruct face from the given skull. These methods use stable diffusion model~\cite{zhao2023uni}, conditioned with text prompts for more controllable and improved face generation. However, these studies are conducted on 3D CT scans of Chinese individuals, which are very difficult to get due to high cost and high radiation exposure. Notably, these diffusion-based models are still under exploration for craniofacial reconstruction.

\section{Dataset}\label{proposed dataset}
\begin{figure*}[t]
        \centering
        \includegraphics[width =\linewidth,keepaspectratio,trim={0cm 3.6cm 0cm 0cm},clip]{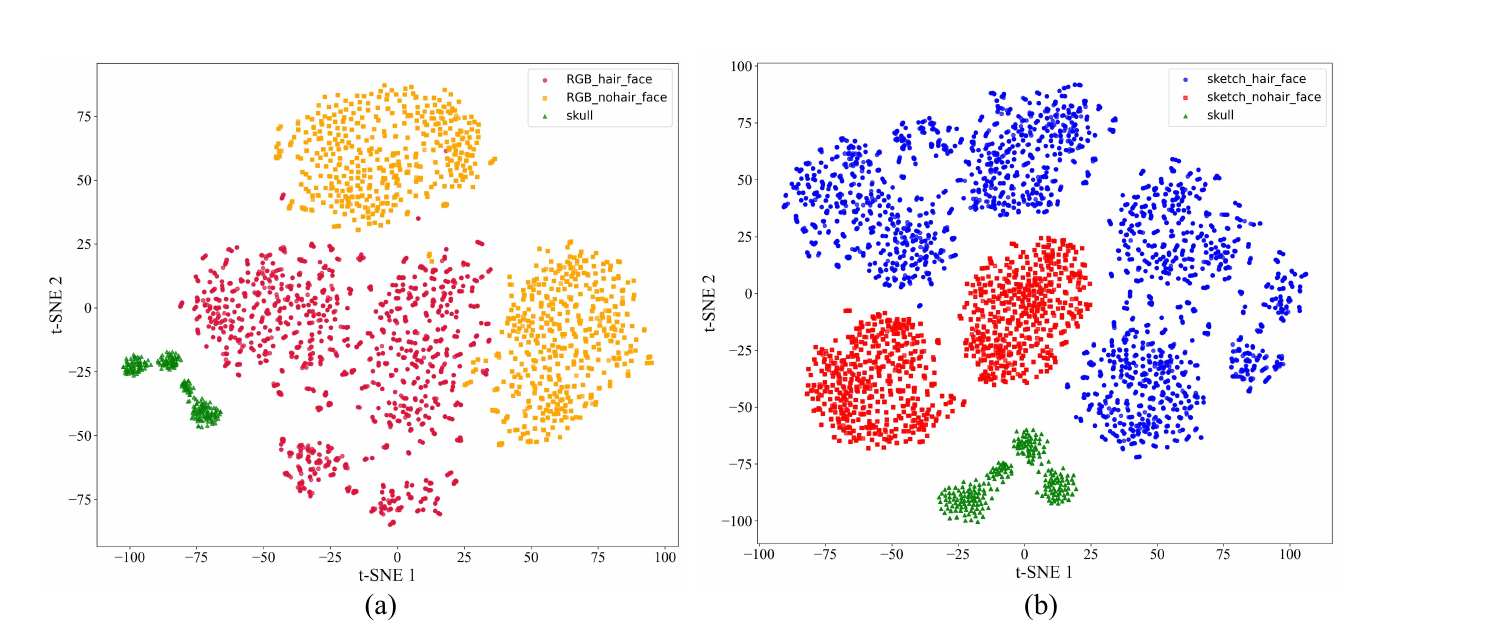}
        \caption{ 2D t-SNE visualization of skull and face embeddings across different facial modalities. (a) shows embeddings of RGB faces with hair and without hair together with skull embeddings, while (b) shows embeddings of sketch faces with hair and without hair together with skull embeddings. Best viewed in color. }
        \label{fig:tsne}
    \end{figure*}

We have obtained S2F dataset as mentioned in~\cite{prasad2025fcr}. We extended the dataset to 120 subjects following the same acquisition procedures and protocols. Required ethical approval has been taken from the institute ethical committee for the collection of required dataset. Also, the required volunteer consent form has been signed by the applicant for contributing to this dataset and research outputs. The proposed Skull-to-Face (S2F) dataset contains paired 2D skull X-ray and facial images captured from both frontal and lateral views. To focus on craniofacial structural information, the facial images were preprocessed to remove hair-related attributes, including hair, beard, and mustache, using the Google Gemini tool~\cite{gemini2024}, thereby retaining only the craniofacial landscape. We use the prompt " age, gender, body mass, face orientation" to model craniofacial variations. Then, the facial images were synthesized across three age groups (25, 45, and 65 years) and three body mass index (BMI) variations (+10\%, baseline, and -10\%). This results in $9 (=3\times3)$ face variations for each view. As we have two views, hence $18 (=2\times9)$ total variations per subject. This process contributes 2,160 $(= 18\times120)$ paired skull-face samples from frontal and lateral views. Furthermore, to increase cross-view skull-face associations, both frontal and lateral skull images were paired with facial images from both views, contributing $36$ skull-face pairs. Finally, we have a total of $4,320 (=36\times120)$ paired samples. To further support multi-modal forensic analysis, the same pairing protocol is applied to four facial modalities: RGB faces, hair-removed faces, sketch faces, and hair-removed sketch faces. To get sketches from the RGB face images, we used the Canny-derived facial edge detector filter~\cite{canny1986computational}. This results in a total of 17,280 multi-modal skull-face associations. Sample images from the prepared dataset are illustrated in Figure~\ref{fig:s2f_dataset} and Figure~\ref{fig:full_s2f}. The dataset was divided into 90\% training and 10\% testing subsets. To further improve the diversity of the training data, nine different data augmentation techniques were applied, increasing the training set to 38,880 paired skull-face samples per modality. Here, we want to clarify that each synthetically generated image is cross-verified and validated by four distinct volunteers to ensure the identity of individuals in the proposed dataset. Additionally, we also visualize the distribution of our curated dataset in 2D t-SNE plot in Figure~\ref{fig:tsne}. 

\section{Methods}\label{methods}

\begin{figure*}[!t]
        \centering
        \includegraphics[width =\textwidth,keepaspectratio,trim={0cm 0cm 0cm 0cm},clip]{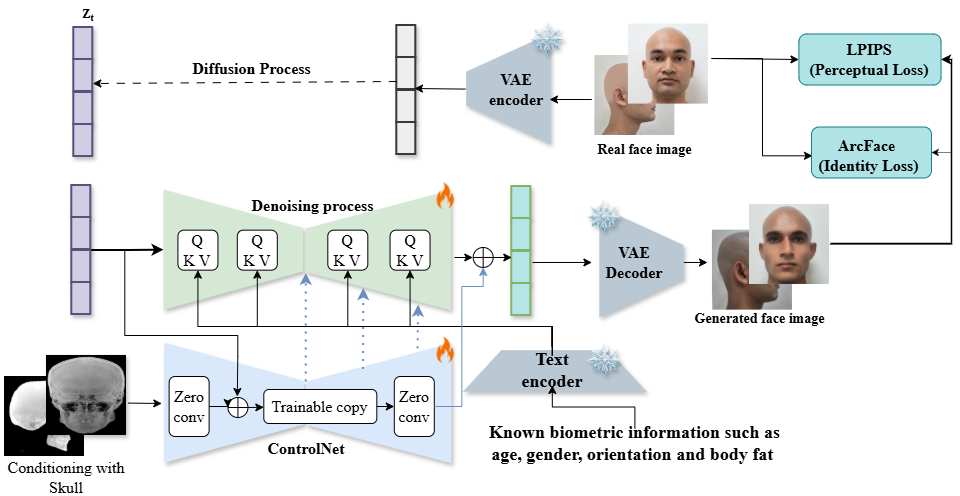}
       \caption{Proposed \textit{Cranio-Diff} framework: In this framework, first a frozen VAE encoder maps the real face image into latent space, where forward diffusion produces a noisy latent $z_t$. A trainable ControlNet branch conditions the denoising UNet on skull morphology and the UNet is jointly guided by biometric text embeddings (age, gender, orientation, body fat) from a frozen text encoder. The denoised latent is decoded by a frozen VAE decoder to synthesize the final face image and is supervised by LPIPS perceptual loss and ArcFace loss against the real face.}
        \label{fig:framework}
    \end{figure*}

In this section, we introduce \textit{Cranio-Diff}, a diffusion-based framework
for skull-conditioned facial reconstruction that integrates structural biometric
conditioning with soft semantic guidance, as depicted in Figure~\ref{fig:framework}.
During training, a pretrained Stable Diffusion model~\cite{rombach2022high} combined
with ControlNet~\cite{zhao2023uni} is employed to establish skull-craniofacial
correspondence and generate structurally consistent facial reconstructions. A frozen
text encoder~\cite{radford2021learning} provides auxiliary semantic control through
standardized biometric text prompts encoding age, gender, pose (i.e., Frontal or Lateral), and body fat (i.e., BMI).
The framework is supervised using perceptual and structural preserving objectives to
ensure visual fidelity and biometric consistency.
% \vspace{-3mm}
\subsection{Overview}

The proposed framework consists of three major components. \textbf{First}, skull-based structural conditioning via ControlNet. \textbf{Second}, biometric text-guided diffusion generation and \textbf{third}, perceptual and structural-preserving refinement.
During training, only the ControlNet branch and the denoising 
UNet are fine-tuned, while the VAE encoder, VAE decoder, and text
encoder remain frozen. This helps in preserving the generative capacity of the pretrained diffusion backbone with minimal training parameters.

\subsection{Skull-Conditioned Structural Conditioning}

Given a skull image $X_s \in \mathbb{R}^{H \times W \times 3}$, the proposed
framework employs a ControlNet branch to extract and inject structural anatomical
conditioning into the denoising UNet. The skull image is first processed through
a zero-initialized convolution layer, ensuring that at initialization the
conditioning signal introduces no perturbation to the pretrained UNet weights:
\begin{equation}
    F_s^{(0)} = \mathcal{Z}_{in}(X_s), \quad F_s^{(0)} \in \mathbb{R}^{C \times H \times W}
\end{equation}
where $\mathcal{Z}_{\text{in}}(\cdot)$ denotes the input zero convolution. The
projected skull features are subsequently combined with intermediate UNet encoder
activations via element-wise addition and propagated through a trainable copy of
the UNet encoder $\mathcal{E}_{\text{ctrl}}(\cdot)$:
\begin{equation}
    F_s = \mathcal{E}_{ctrl}(F_s^{(0)} \oplus F_{unet}), \quad F_s \in \mathbb{R}^{C \times H \times W}
\end{equation}
Here, $F_{unet}$ denotes the feature map extracted from the encoder block of the denoising UNet. A second zero-initialized convolution $\mathcal{Z}_{\text{out}}(\cdot)$ then
projects these features they are injected into the corresponding UNet blocks through residual feature fusion.
\begin{equation}
    C_s = \mathcal{Z}_{out}(F_s), \quad C_s \in \mathbb{R}^{C \times H \times W}
\end{equation}
This design allows the ControlNet to learn rich skull-to-face structural mappings
while leaving the pretrained generative backbone intact. The zero convolution
initialization ensures training stability by preventing early-stage noise
propagation into the denoising UNet.
\subsection{Biometric Text Conditioning}

To incorporate soft semantic priors alongside structural skull conditioning, known
biometric attributes, which include subject age, gender, head orientation, and body
fat index. These attributes are encoded as standardized textual descriptions. Given a text prompt
$\mathcal{T}$ encoding these attributes, the frozen text encoder $E_\tau(\cdot)$
produces semantic embeddings:
\begin{equation}
    \tau = E_\tau(\mathcal{T}), \quad \tau \in \mathbb{R}^{L \times d}
\end{equation}
where $L$ denotes the token sequence length and $d$ the embedding dimension. These
embeddings are injected into each cross-attention block of the denoising UNet as
keys and values, while queries are derived from intermediate noisy latent features.
The cross-attention operation is formulated as:
\begin{equation}
    \text{Attention}(Q, K, V) = \text{softmax}\left(\frac{QK^\top}{\sqrt{d}}\right)V
\end{equation}
where $Q = z_t W_Q$, $K = \tau W_K$, $V = \tau W_V$, and $W_Q, W_K, W_V \in
\mathbb{R}^{d \times d}$ are learnable projection matrices within the denoising UNet. This mechanism enables the model to modulate facial generation according to
biometric priors, complementing the hard structural constraints provided by the
ControlNet branch.
\subsection{Skull-conditioned Diffusion Generation}

Given a real facial image $x_0$, the VAE encoder $\mathcal{E}(\cdot)$ maps it into
a compact latent representation $z_0$ . The forward diffusion process progressively
corrupts this latent with Gaussian noise over $T$ timesteps:
\begin{equation}
    q(z_t | z_{t-1}) = \mathcal{N}(z_t; \sqrt{1 - \beta_t}\, z_{t-1},\, \beta_t I)
\end{equation}
At timestep $t$, the noisy latent is expressed as:
\begin{equation}
    z_t = \sqrt{\bar{\alpha}_t}\, z_0 + \sqrt{1 - \bar{\alpha}_t}\, \epsilon
\end{equation}
where $\epsilon \sim \mathcal{N}(0, I)$ and $\bar{\alpha}_t = \prod_{s=1}^{t}(1 -
\beta_s)$. The denoising UNet, jointly conditioned on the skull conditioning
map $C_s$ and text embeddings $\tau$, predicts the injected noise as
$\epsilon_\theta(z_t, t, C_s, \tau)$. The diffusion training objective is:
\begin{equation}
    \mathcal{L}_{\text{diff}} = \mathbb{E}_{z_0, \epsilon, t}
    \left[ \| \epsilon - \epsilon_\theta(z_t, t, C_s, \tau) \|_2^2 \right]
\end{equation}
The denoised latent is subsequently decoded by the frozen VAE decoder
$\mathcal{D}(\cdot)$ to produce the synthesized facial image
$\hat{x}_0 = \mathcal{D}(\hat{z}_0)$. Here, $\hat{z}_0$ is the latent representation estimated by UNet.
\subsection{Structural-Preserving Refinement}

Although the diffusion objective ensures structural skull correspondence, additional
supervision is required to preserve perceptual quality and biometric structural consistency in the generated facial images. Accordingly, two complementary loss
functions are employed.

\textbf{Perceptual Loss.} To improve perceptual fidelity between the generated
and real facial images, LPIPS perceptual loss~\cite{zhang2018unreasonable} is applied:
\begin{equation}
    \mathcal{L}_{\text{LPIPS}} = \sum_l \| \phi_l(\hat{x}_0) - \phi_l(x_0) \|_2^2
\end{equation}
where $\hat{x}_0$ denotes the generated facial reconstruction and $\phi_l(\cdot)$
represents deep feature activations extracted from a pretrained perceptual network
at layer $l$.

\textbf{Structural Loss.} To enforce biometric structural consistency, the generated
image is projected into the ArcFace embedding space. The embeddings of the
generated and real facial images are computed as $z_{\text{gen}} = F_{\text{arc}}
(\hat{x}_0)$ and $z_{\text{gt}} = F_{\text{arc}}(x_0)$, where $z_{gen}, z_{gt} \in \mathbb{R}^{d}$ respectively, and the
structural consistency loss~\cite{deng2019arcface} is formulated as:
\begin{equation}
    \mathcal{L}_{\text{id}} = 1 - \frac{z_{\text{gen}}^\top z_{\text{gt}}}
    {\| z_{\text{gen}} \| \| z_{\text{gt}} \|}
\end{equation}

\textbf{Total Objective.} The final optimization objective jointly minimizes
structural, perceptual, and structural-preserving losses:
\begin{equation}
    \mathcal{L} = \mathcal{L}_{\text{diff}} + \lambda_1 \mathcal{L}_{\text{LPIPS}}
    + \lambda_2 \mathcal{L}_{\text{id}}
\end{equation}

where $\lambda_1$ (= 0.20) and $\lambda_2$ (= 0.20) control the relative contributions of perceptual
and structural-preserving regularization, respectively. The values of $\lambda_1$ and $\lambda_2$ are selected empirically. During optimization, gradients are propagated through both the ControlNet branch and the denoising UNet, ensuring training efficiency and stability.
\section{Experiments}\label{experiments}
The primary objective of forensic craniofacial reconstruction is to identify unknown deceased individuals; thus, the effectiveness of reconstructed faces in recognition algorithms is essential. Hence, we tested various baselines with benchmark S2F dataset with different evaluation metrics. We also evaluate the retrieval performance of our proposed model with various face recognition backbones on different retrieval metrics.
\subsection{Implementation Details}
The proposed  \textit{Cranio-Diff} method is implemented in Pytorch and trained on Nvidia H200 GPU with 141 GB VRAM, designed for complex deep learning computations. We trained our proposed framework for 350 epochs. We utilized Realistic Vision v5.1 (fine-tuned version of Stable Diffusion v1.5), as the backbone of generative framework. All images are resized to 512 x 512 spatial resolution. We selected the best hyperparameters empirically: a learning rate of 0.0001, a batch size of 14, a weight decay of 0.00001 using the AdamW optimizer. For training our proposed model with S2F dataset, we partitioned this dataset at subject level (to prevent identity leakage) into training (108 subjects) and test (12 subjects) with the split ratio of 90:10 respectively. Total nine types of data augmentation techniques are applied to improve the generalization and robustness of the proposed method, which includes contrast jitter ($\pm20\%$, $0.8{-}1.2$), brightness jitter ($\pm15\%$, $0.85{-}1.15$), horizontal flipping (probability = $1.0$), small-angle rotation with Gaussian noise ($\pm5^\circ$, $\sigma=0.01$), and random erasing (maximum erased area = $5\%$). Furthermore, composite augmentations were also employed, including contrast + rotation ($(0.8{-}1.2)$ + $\pm5^\circ$), brightness + noise ($(0.85{-}1.15)$ + $\sigma=0.01$), flip + contrast (flip + $(0.8{-}1.2)$), and erase + brightness + rotation (combined augmentation). All frontal and lateral views, synthetic age and BMI variations, and various augmentations are derived within that subject's split and all results are reported based on the test set.

\subsection{Experimental results}
\textbf{Quantitative comparison.} The proposed framework is evaluated on S2F dataset, we have used Fréchet Inception Distance (FID)~\cite{heusel2017gans}, Structural Similarity Index scores (SSIM)~\cite{wang2004image}, LPIPS~\cite{zhang2018unreasonable}. Table ~\ref{tab:different_modalities} provides quantitative results on different baseline methods and our model achieves the improved performance across majority of mentioned metrics, with FID of 58.89, SSIM of 0.82, LPIPS of 0.24, which indicates that the generated face has improved photorealistic appearance and semantically alignment with its corresponding skull. We have also evaluated retrieval performance using three backbones (i.e., FaceNet, ArcFace and VGGFace) using three retrieval metrics (i.e, recall@k, mAP@k and MRR@k) with different gallery sizes, as shown in Table~\ref{tab:retrieval_results}. From this table, we have observed that out of three backbones, VGGFace is performing better with recall@10 of 69.23, mAP@10 of 17.93, MRR@10 of 32.00 on gallery size of 100. It is also observed that when gallery size increase from 100 to 200, there is small reduction in the values of mentioned retrieval metrics. 
\begin{table*}[!ht]
\centering
\caption{Quantitative comparison of different methods on S2F dataset. Here, FID, LPIPS and SSIM scores are reported.}
\label{tab:different_modalities}
\resizebox{\linewidth}{!}{
\begin{tabular}{l|ccc|ccc|ccc|ccc}
\toprule
\multirow{2}{*}{\textbf{Method}}
&
\multicolumn{3}{c|}{\textbf{RGB Face}}
&
\multicolumn{3}{c|}{\textbf{Hair-Removed Face}}
&
\multicolumn{3}{c|}{\textbf{Sketch Face}}
&
\multicolumn{3}{c}{\textbf{Hair-Removed Sketch}}
\\

\cmidrule(lr){2-4}
\cmidrule(lr){5-7}
\cmidrule(lr){8-10}
\cmidrule(lr){11-13}

&
FID$\downarrow$ &LPIPS$\downarrow$ & SSIM$\uparrow$
&
FID$\downarrow$ &LPIPS$\downarrow$ & SSIM$\uparrow$
&
FID$\downarrow$ &LPIPS$\downarrow$ & SSIM$\uparrow$
&
FID$\downarrow$ &LPIPS$\downarrow$ & SSIM$\uparrow$
\\
\midrule

CycleGAN~\cite{zhu2017unpaired} &206.91
 &0.53 &\textbf{0.57}&93.82
 &0.48 &0.60&293.82
 &0.68 &\textbf{0.70}&379.64
 &0.62&0.63 \\

Pix2Pix~\cite{isola2017image} &400.00
 & 0.74&0.11&178.62
 &0.47 &0.70&285.22
 & 0.66&0.68&294.48
 &0.65 &\textbf{0.69}\\

CUT~\cite{park2020contrastive} &203.28
 & 0.52&0.53&86.42
 & 0.45&0.64&326.59
 & 0.52&0.47&336.33
 &0.60&0.51 \\

ControlNet~\cite{zhao2023uni} &90.55
 & 0.59&0.30&107.82
 & 0.43&0.49&182.11
 &0.75 &0.16&110.21
 &0.42&0.60 \\

 ICCR-Diff~\cite{zhang2025iccr}&77.13
 & 0.46&0.45&72.97
 &0.40 &0.56&\textbf{74.21}
 &0.50 &0.23&101.18
 &0.40&0.50 \\
 \textbf{Cranio-Diff (Ours)} &\textbf{60.51}
 & \textbf{0.39}&0.56&\textbf{58.89}
 &\textbf{0.24} &\textbf{0.82}&80.45
 &\textbf{0.48} &0.19&\textbf{64.47}
 &\textbf{0.26}&0.45 \\

\bottomrule
\end{tabular}
}
\end{table*}

% \begin{table*}[!ht]
% \centering
% \caption{Quantitative comparison of different methods for skull-to-face generation. Lower FID and LPIPS indicate better perceptual quality, while higher IS, SSIM, PSNR, and ArcFace indicate better realism and identity preservation.}
% \label{tab:all_metrics_comparison}

% \resizebox{\columnwidth}{!}{%
% \begin{tabular}{l|cccccc}
% \toprule

% \textbf{Methods}  
% & \textbf{FID}$\downarrow$ 
% & \textbf{IS}$\uparrow$ 
% & \textbf{SSIM}$\uparrow$ 
% & \textbf{LPIPS}$\downarrow$ 
% & \textbf{PSNR}$\uparrow$
% & \textbf{ArcFace}$\uparrow$ \\

% \midrule

% CycleGAN~\cite{zhu2017unpaired} 
% & 93.82 & 1.18 & 0.60 & 0.48 & 15.38 & 0.14 \\

% Pix2Pix~\cite{isola2017image} 
% & 178.62 & 1.32 & 0.70 & 0.47 & 16.27 & 0.13 \\

% ICCR-Diff~\cite{zhang2025iccr} 
% & 72.97 & \textbf{1.67} & 0.56 & 0.40 & 14.78 & 0.23 \\

% \textbf{Cranio-Diff (Ours)} 
% & \textbf{58.89} & 1.52 & \textbf{0.82} 
% & \textbf{0.24} & \textbf{17.21} & \textbf{0.25} \\

% \bottomrule
% \end{tabular}
% }
% \end{table*}

\begin{table*}[!ht]
    \centering
     \caption{Ablation studies on Loss functions and text prompts. Here, w/o means without.}
     %\resizebox{\linewidth}{!}{%
    \begin{tabular}{l|ccc}
    \toprule
    \textbf{Configuration}
    & \textbf{FID}$\downarrow$
   
    & \textbf{SSIM}$\uparrow$
    & \textbf{LPIPS}$\downarrow$\\

    \midrule
    
    Cranio-Diff (w/o text prompts) &64.46&0.78&0.31\\
 Cranio-Diff (w/o ArcFace loss) &64.29&0.80&0.28\\
    Cranio-Diff (w/o LPIPS loss) &\textbf{53.28}&0.81&0.25\\
    \textbf{Cranio-Diff ( with text+both losses)}  & 58.89 & \textbf{0.82} 
& \textbf{0.24} \\
    
    \bottomrule
    \end{tabular}

    \label{tab:loss-prompt}
\end{table*}

\begin{table*}[!ht]
\centering
\caption{Retrieval performance comparison under different gallery sizes on the S2F dataset. Higher values indicate better retrieval performance.}
\label{tab:retrieval_results}

\resizebox{\textwidth}{!}{
\begin{tabular}{l|l|cccccc|cccccc}
\toprule

\multirow{3}{*}{\textbf{Method}}
& \multirow{3}{*}{\textbf{Backbone}}

& \multicolumn{6}{c|}{\textbf{Gallery Size = 100}}
& \multicolumn{6}{c}{\textbf{Gallery Size = 200}} \\

\cmidrule(lr){3-8}
\cmidrule(lr){9-14}

&
& \multicolumn{2}{c}{Recall}
& \multicolumn{2}{c}{mAP}
& \multicolumn{2}{c|}{MRR}

& \multicolumn{2}{c}{Recall}
& \multicolumn{2}{c}{mAP}
& \multicolumn{2}{c}{MRR} \\

\cmidrule(lr){3-8}
\cmidrule(lr){9-14}
&
& @10 & @25
& @10 & @25
& @10 & @25

& @10 & @25
& @10 & @25
& @10 & @25 \\

\midrule

\multirow{3}{*}{ICCR-Diff~\cite{zhang2025iccr}}

&FaceNet~\cite{schroff2015facenet}
& 36.67 & 61.67 & 5.71 & 8.38 & 11.55 & 13.08
& 21.39 & 41.67 & 2.78 & 4.33 & 6.19 & 7.48 \\

& ArcFace~\cite{deng2019arcface}
& 41.11 & 66.94 & 9.54 & 13.35 & 15.76 & 17.36
& 27.78 & 47.58 & 4.85 & 6.91 & 8.08 & 9.32 \\

& VGGFace~\cite{parkhi2015deep}
& 30.28 & 51.11 & 5.52 & 8.36 & 9.82 & 11.12
& 17.78 & 35.56 & 2.92 & 4.39 & 5.55 & 6.62 \\

\midrule

\multirow{3}{*}{\textbf{Cranio-Diff (Ours)}}

 &FaceNet~\cite{schroff2015facenet}
& 61.97 &86.75 & 11.54 & 14.23 & 23.20 & 23.50
& 39.74 & 67.95 & 6.46 & 8.31 & 13.72 & 15.52 \\

& ArcFace~\cite{deng2019arcface}
& 64.53 & \textbf{98.17} & 13.54 & 16.35 & 28.49 & 28.74
& 46.15 & 72.65 & 7.33 & 9.55 & 17.08 & 18.78 \\

& \textbf{VGGFace}~\cite{parkhi2015deep}
& \textbf{69.23} & 91.88 & \textbf{17.93} & \textbf{21.25} & \textbf{32.00} & \textbf{32.20}
& \textbf{48.72} & \textbf{72.65} & \textbf{10.76} & \textbf{13.11} & \textbf{20.36} & \textbf{21.94} \\

\bottomrule
\end{tabular}
}
\end{table*}

\begin{figure*}[!ht]
        \centering
        \includegraphics[width =\linewidth,keepaspectratio,trim={3cm 0.5cm 0cm 1cm},clip]{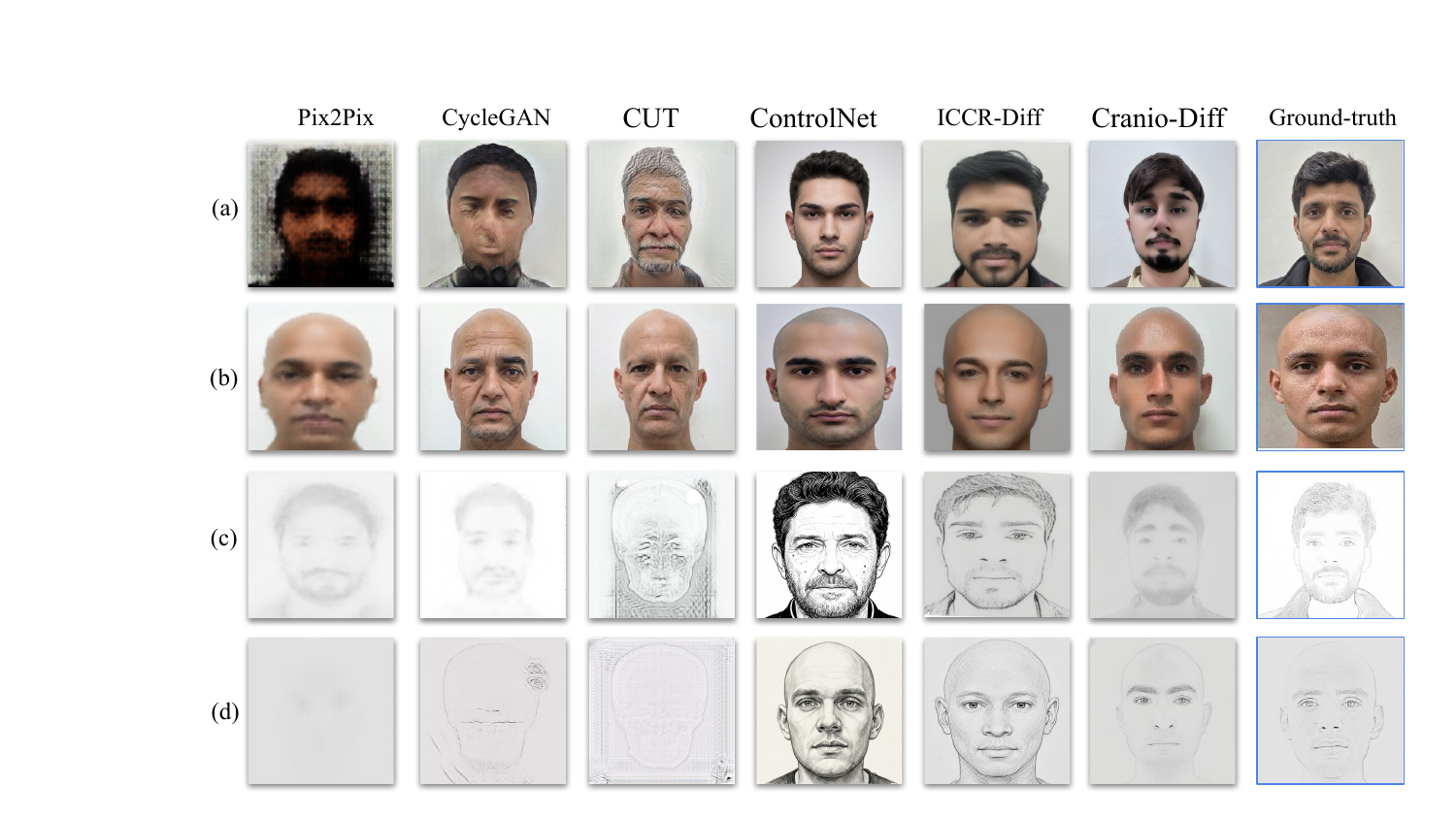}
        \caption{ Qualitative results on generated images with ground truth images across different modalities for different generative methods. Here (a), (b), (c) and (d) show results on different face modalities.}
        \label{fig:all_meth_gen}
    \end{figure*}
    
\begin{figure*}[!ht]
        \centering
        \includegraphics[width =\linewidth,keepaspectratio,trim={2cm 0.6cm 2cm 2cm},clip]{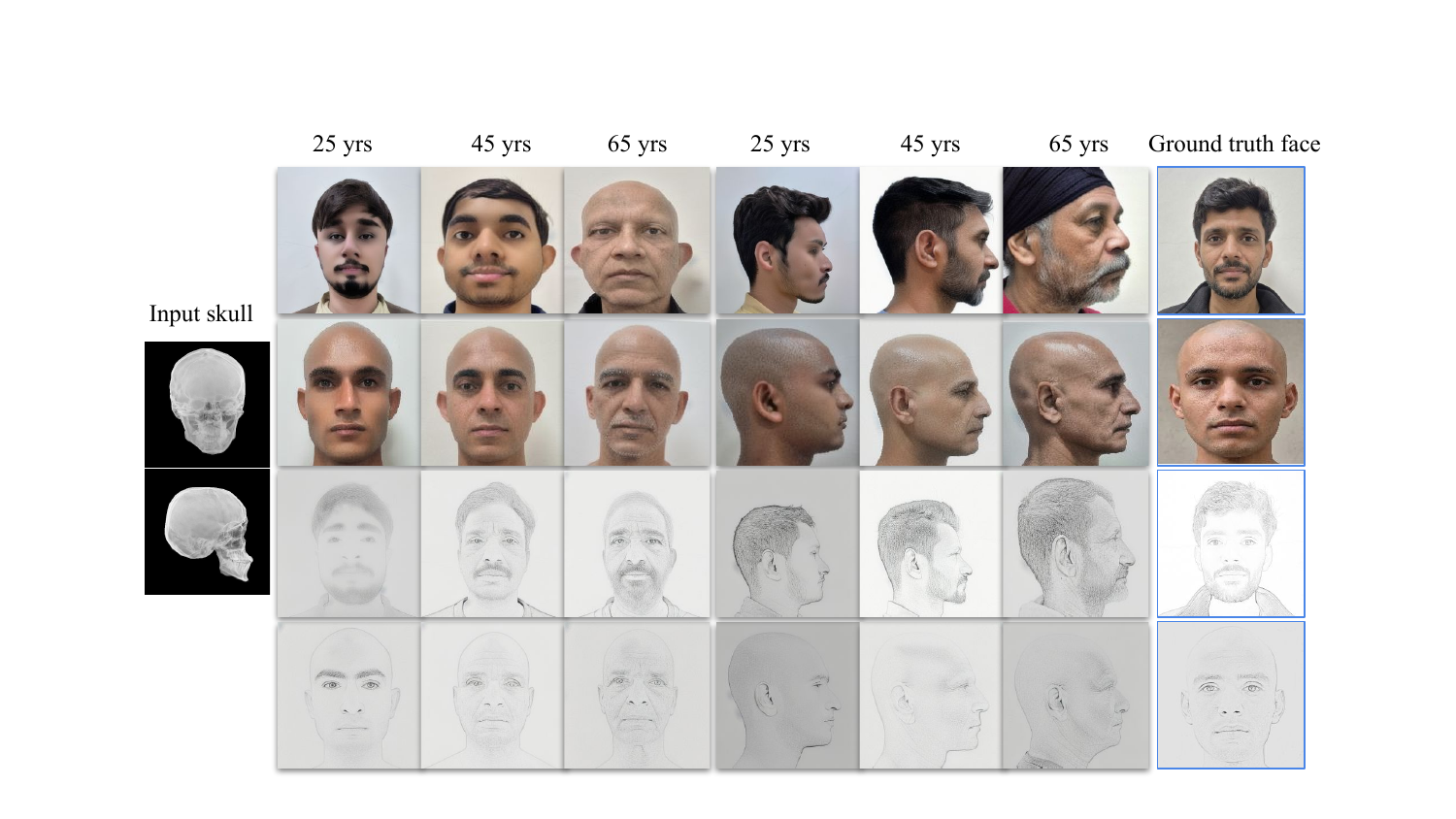}
        \caption{ Qualitative results on generated images with ground truth images across different modalities for age variations (25 yrs, 45 yrs and 65 yrs) of Cranio-Diff method when input skull image is given. }
        \label{fig:craniodiff_age}
    \end{figure*}

\begin{figure*}[!ht]
        \centering
        \includegraphics[width =\linewidth,keepaspectratio,trim={2cm 0.7cm 2.5cm 2cm},clip]{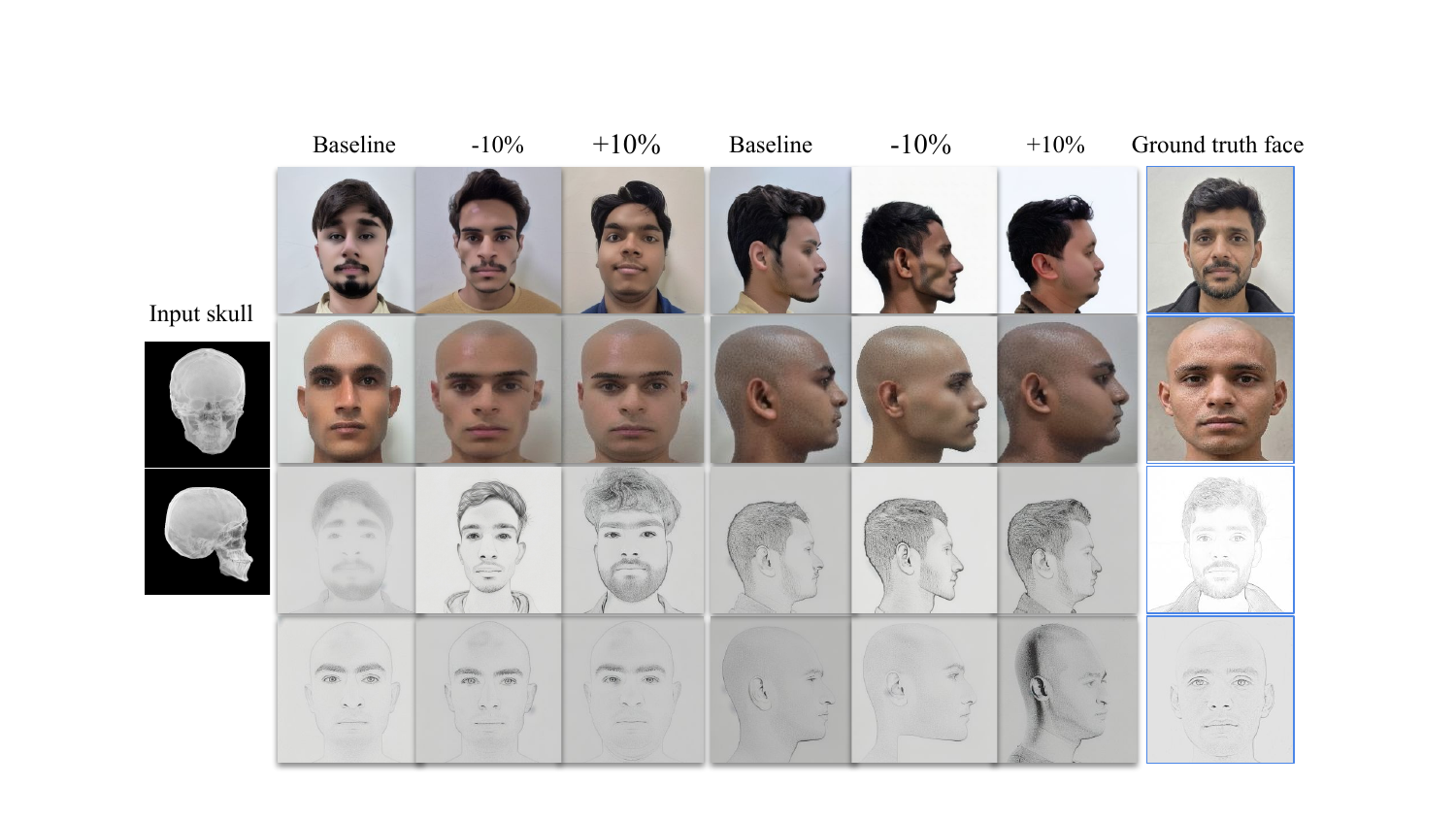}
        \caption{ Qualitative results on generated images with ground truth images across different modalities for BMI variations (-10\%, baseline and +10\%) of Cranio-Diff method when input skull image is given.  }
        \label{fig:craniodiff_bmi}
    \end{figure*}
\begin{figure}
    \centering
    
    \includegraphics[width=\linewidth,keepaspectratio,trim={0cm 1cm 0cm 1cm},clip]{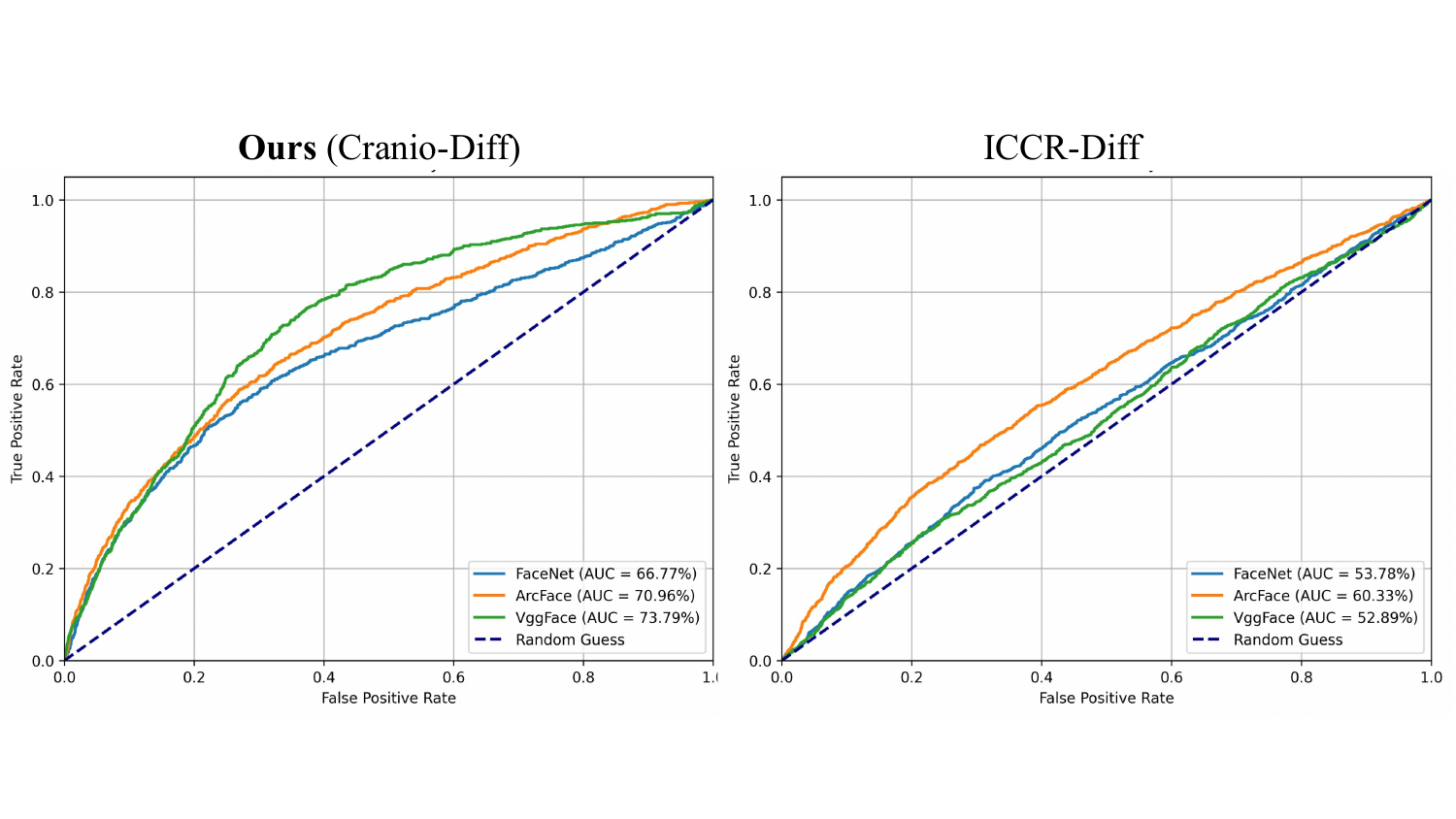}
    
    \caption{Qualitative results showing ROC-AUC curve with gallery size of 100 on S2F dataset. Here left curve represents ours (\textit{Cranio-Diff}) while right one represents ICCR-Diff. Best viewed in colors.}
    \label{fig:q3results}
\end{figure}
\begin{figure*}[!t]
        \centering
        \includegraphics[width =\linewidth,keepaspectratio,trim={1.5cm 2.1cm 0cm 2cm},clip]{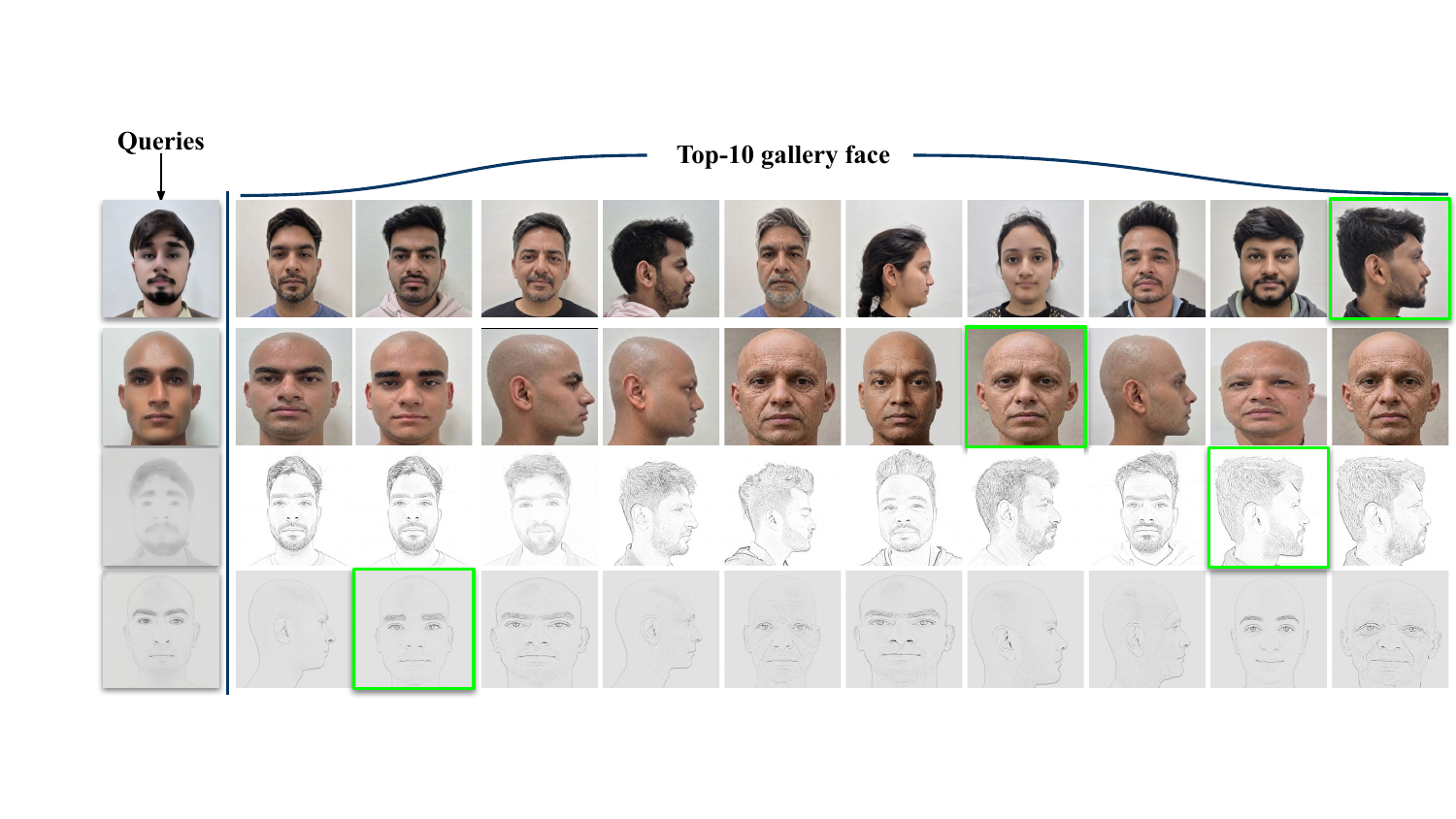}
        \caption{ Qualitative results on top-10 retrieval across different modalities. Here, queries are the generated image from different modalities and face image with green box are the ground truth images. }
        \label{fig:retrieval}
\end{figure*}

\textbf{Qualitative comparison.}
Figure~\ref{fig:all_meth_gen} shows the qualitative comparisons between different generative models with our proposed framework. From this figure, it is observed that our FCR method reconstructs photorealistic face images. To show the effectiveness of \textit{Cranio-Diff} across different craniofacial variations, Figure~\ref{fig:craniodiff_age} and Figure~\ref{fig:craniodiff_bmi} show results on generated face images with different variations of age and BMI with high fidelity and semantic details. Figure~\ref{fig:q3results} illustrates ROC-AUC curve comparison with three different backbones on S2F dataset with the gallery size of 100. From this ROC-AUC curve, it is observed that our proposed method \textit{Cranio-Diff} shows promising results compared to the state-of-the-art method (i.e., ICCR-Diff) with the best backbone of VGGFace, achieving an ROC-AUC of 73.79\%. Figure~\ref{fig:retrieval} shows qualitative results on retrieval of top-10 faces from the gallery for the given generated face images across different facial modalities. For retrieval of faces, we have utilized our best performing backbone, which is VGGFace. Here, the similarity of the generated face with the faces in the gallery is calculated on the basis of the Euclidean distance between the embeddings of images. For retrieval, the gallery contains multiple images of the correct identity for the given query image.   
% \begin{figure}
%     \centering
    
%     \includegraphics[width=\linewidth,keepaspectratio,trim={5cm 4cm 5cm 2cm},clip]{Figure/cranio-diff_retrieval.pdf}
    
%     \caption{Qualitative results on retrieval of top-10 faces from the gallery for given generated faces as query (in orange box) in frontal and lateral views. Here, the green box represents a correct match for the given query. For a given query multiple faces are present in the gallery with variation in age and BMI. Best viewed in colors.}
%     \label{fig:q4results}
% \end{figure}

\subsection{Ablation Studies}
To evaluate the contribution of each component in the proposed framework, we conduct ablation studies on both biometric text conditioning and loss functions. The effect of biometric text prompts and loss functions is analyzed in Table~\ref{tab:loss-prompt}. For loss ablation, two variants are considered: (i) removing the ArcFace loss while retaining the LPIPS perceptual loss, and (ii) removing the LPIPS loss while retaining the ArcFace loss. The results demonstrate that both losses contribute positively to the reconstruction quality and structural preservation of the generated facial images. When both losses are jointly employed, the proposed \textit{Cranio-Diff} framework achieves the best overall performance.
\subsection{Limitations and Discussion}
In forensic craniofacial reconstruction, due to absence of public benchmark dataset, this study is conducted on curated S2F dataset, which consists of only Indian individuals. Hence, due to this, model may capture and learn population specific features which may limits its generalizability across different ethnicity. Also, when comes to retrieving faces from the gallery database for a given generated query face image, it is observed that the reported results at different values of k and gallery size have high recall@k but low mAP@k. This indicates that when correct identity is retrieved, it is not consistently ranked at the top positions. Hence, future work could be in the direction of collecting more diverse dataset, leveraging multi-modalities and in building a methods for better feature disentanglement and representation learning. We want to emphasize that the reconstructed face image is a probabilistic approximation based on specific population data. Hence, the proposed \textit{Cranio-Diff} framework can be seen as a scientific tool to assist forensic experts in craniofacial identification and anthropological investigations. However, it should not be considered as a replacement for established standard forensic identification methods rather, it may provide complementary evidence to assist expert assessment and guide further investigations.\section{Conclusion}\label{conclusion}
This work presents a unified framework for craniofacial reconstruction using 2D frontal and lateral X-ray facial scans. It is easy to collect 2D X-ray-face data since getting a face X-ray scan is more affordable and convenient than other methods. Our proposed approach leverages ControlNet guided Stable Diffusion model (i.e, Realistic Vision v5.1) with dual-view X-ray conditioned map to preserve fine grained structural information. With more diverse face dataset across age and BMI, enables the model to learn variation in face across different age and body mass. Additionally, integrating corresponding text prompt via CLIP text encoder helps the model to incorporates semantic details. By comparative quantitative and qualitative analysis of different generative models for skull to face image translation problem, we have found that our proposed framework shows improved performance over the other generative AI frameworks in cross-domain image translation which is generating realistic human faces from the given 2D X-ray skull images.

\vspace{0.5cm}
\textbf{Data availability}
 Due to ethical concerns, our skull-face data will be made available on request with a proper data use agreement. The dataset used in this study, which includes X-ray images and their corresponding face vectors (vector representation of face images), is made publicly available at: \url{https://github.com/singh-ml/IIT_Mandi_S2F}.
 However, the original faces with their corresponding skull images cannot be shared due to privacy and ethical concerns.
\vspace{0.5cm}

\textbf{Code availability}
 The code used in this study will be released once the manuscript is accepted.
 \vspace{0.5cm}
 
\textbf{Funding} This research is funded by seed grant project No. IITM/SG/DIS-ROS-SPA/111, Indian Institute of Technology Mandi, Department of Higher Education, Ministry of Education, Government of India.
\vspace{0.5cm}

\textbf{Author Contribution declaration}
Conceptualization, Ravi Shankar Prasad, Naresh Gurjar; Methodology, Ravi Shankar Prasad, Naresh Gurjar, Shashank Baghel and Dinesh Singh; Data curation, Ravi Shankar Prasad, Naresh Gurjar, Chirag; Original Draft Preparation, Ravi Shankar Prasad; Review \& Editing, Ravi Shankar Prasad and Dinesh Singh.

\section{Declaration}\label{declaration}

\textbf{Ethical approval} This study has been reviewed and approved by the Institute Ethical Committee, REF: IITM/IEC(H)/2024/DS/P1, IIT Mandi.
\vspace{0.2cm}

\paragraph{\textbf{Consent to participate}} Informed consent for participation was obtained from all the subjects who are involved in this study.
\vspace{0.2cm}

\paragraph{\textbf{Consent for publication}} The data used in this study were anonymized properly and do not contain personal information for subject identification.
\vspace{0.2cm}

\paragraph{\textbf {Competing interests}} The authors declare that they have no competing interests. 

\section{Acknowledgment}\label{anknowledgment}
 For creating this dataset, we sincerely want to thank the volunteers who dedicated their time and effort to help create the dataset called \emph{S2F}. We also appreciate the cooperation and assistance from the Institute Health Centre during the data collection process and IIT Mandi for necessary support to this research. During the preparation of this work, the author(s) used Grammarly in order to correct the spelling and grammar. After using this tool, the author(s) reviewed and edited the content as needed and take(s) full responsibility for the content of the publication.

%\bibliographystyle{cas-model2-names}

\bibliography{references}
\end{document}